# A New Dataset and Proposed Convolutional Neural Network Architecture for Classification of American Sign Language Digits

Arda Mavi

Ayrancı Anadolu High School, Ankara, Turkey

**Abstract**

According to interviews with people who work with speech impaired persons, speech impaired people have difficulties in communicating with other people around them who do not know the sign language, and this situation may cause them to isolate themselves from society and lose their sense of independence. With this paper, to increase the quality of life of individuals with facilitating communication between individuals who use sign language and who do not know this language, a new American Sign Language (ASL) digits dataset that can help to create machine learning algorithms which need to large and varied data to be successful created and published as "Sign Language Digits Dataset" on Kaggle Datasets web page, a proposal Convolutional Neural Network (CNN) architecture that can get 98% test accuracy on our dataset presented, and compared with the existing popular CNN models. Dataset available at https://www.kaggle.com/ardamavi/sign-language-digits-dataset .

## 1 Introduction

Sign language is one of the popular methods used by speech and hearing impaired people and people around them to communicate with each other, ASL is used by approximately 250,000-500,000 Americans (and some Canadians) [1]. Artificial intelligence technologies, which can provide oral communication between different languages by converting speech to text and translating language, are popularly used in people's daily lives but speech and hearing impaired people can not use these technologies when communicating with people around them who do not know the sign language because the sign language consists of certain hand gestures. One of the methods that can be used to translate sign language can be done by classification of images that contain hand movements. With today's powerful computing hardware, the CNN algorithms [2] have achieved remarkable success in real time image classification [3]. With this paper; a new dataset prepared that consisting of digits in ASL, which has the diversity and correct labeled information that CNN architectures need, published with publicly accessible for research projects and to shed light on future works, a CNN architecture that reached 97% success rate has been prepared on our dataset and compared with other CNN architectures.

## 2 Proposed Dataset

### 2.1 Preparation of the Dataset

During the preparations of this study, ASL was chosen because of the high number of users among different sign languages [1]. After the movements of the signs to be collected were determined, 218 volunteer individuals were asked to perform the movements of 10 different determined ASL numbers signs with their right hands and 3024x3024 pixel colored (3-channel RGB) photos were taken in white background with the hands of the volunteers in the middle of the photo with sufficient white light. The collected images was converted to single channel grayscale images with a size of 64x64 pixels and pixel values ranging from 0 to 255 were normalized to 0-1. With this conversion, the images can be processed faster by CNN algorithms without losing data. Sample images from the dataset are shown in Figure 1.

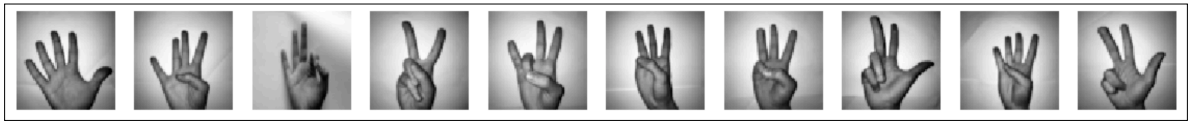

Fig. 1. Sample Images From Our Dataset

### 2.2 Published Dataset

The prepared images were converted to Numpy arrays in the Numpy library in Python language and saved with the file name "X.npy". After applying One Hot Encoding to the digits labels of the visual data, obtained binary class matrices saved to "Y.npy" file as Numpy array.

This two ".npy" extended file published as "Sign Language Digits Dataset" on Kaggle Datasets web page at https://www.kaggle.com/ardamavi/sign-language-digits-dataset .

## 3 Preparation of CNN Architectures

### 3.1 Popular CNN Architectures

In this section, two popular CNN architectures are selected for to use with the dataset, the structure of these architectures and a proposed CNN architecture that created to solve our dataset are introduced.

#### 3.1.1 MVGG-5

MVGG-5 [4] architecture consists of two Convolution (16 filters with 3x3 shape), a Max Pooling (2x2 filter), a Convolution (48 filters with 3x3 shape), a Max Pooling (2x2 filter), a Dense (128 neuron) and another Dense (10 neuron) layer that finished with Softmax activation function. The neurons at the end of the architecture represent classes in the dataset. All output values of every Convolution and Dense (fully connected) layer except last Dense layer are normalized by using ReLu activation function and all Convolution layer have 1x1 and all Pooling layer have 2x2 stride rate.

### 3.1.2 MVGG-9

MVGG-9 [4] architecture consists of two Convolution (16 filters with 3x3 shape), a Max Pooling (2x2 filter), two Convolution (32 filters with 3x3 shape), a Max Pooling (2x2 filter), two Convolution (48 filters with 3x3 shape), a Max Pooling (2x2 filter), a Convolution (64 filters with 3x3 shape), a Max Pooling (4x4 filter), a Dense (128 neuron) and another Dense (10 neuron) layer that finished with Softmax activation function. All output values of every Convolution and Dense layer except last Dense layer are normalized by using ReLu activation function and all Convolution layer have 1x1 and all Pooling layer have 2x2 stride rate except last Max Pooling layer, it has 4x4 strife rate.

MVGG-9 has more layers than MVGG-5, so it requires more processing capacity than MVGG-5.

### 3.2 Proposed Architecture

Proposed CNN architecture consists of a Convolution (32 filters with 3x3 shape), a Convolution (64 filters with 3x3 shape), a Max Pooling (2x2 filter), a Convolution (64 filters with 3x3 shape), a Max Pooling (2x2 filter), a Convolution (128 filters with 3x3 shape), a Max Pooling (2x2 filter), a Dense (526 neuron), a Dense (128 neuron) and another Dense (10 neuron) layer that finished with Softmax activation function. Output values are normalized using the ReLu activation function at the end of each convolution and dense layer. All output values of every Convolution and Dense layer except last Dense layer are normalized by using ReLu activation function and all Convolution layer have 1x1 and all Pooling layer have 2x2 stride rate.

Proposed architecture has more convolution layers than MVGG-5 and less than MVGG-9 architecture, but unlike two MVGG models [4] prepared architecture has an extra dense layer in the classification part.

### 3.3 Train Procedure of Models

80% of the dataset is reserved to be used in the training process of the model, the remaining 20% is reserved as test dataset. The split test data have same number of samples per class and samples were selected randomly.

During the training of models, Dropout layers [3] with 50% ratio were used after the dense layers except for the last layers that hold classes and models are trained using AdaDelta optimization function [5] and Categorical Cross Entropy Loss function [3].

Prepared codes are available at https://github.com/ardamavi/Vocalize-Sign-Language .

## 4 Result and Comparison

With this study, the ASL digit dataset, that can be used in artificial intelligence researches, has been prepared and published with public access on Kaggle. The published dataset was downloaded 14,927 times between December 2017 to November 2020.

CNN algorithms, which are one of the popular algorithms for image classification and can be used to classify images with sign language, were tested with the prepared dataset.

At the end of 15 training epochs, 95% success rate was reached on the test dataset with MVGG-5 , 96% with MVGG-9, and 97% with the proposed architecture. These two selected MVGG architectures are the two most successful architectures on our dataset among MVGG architectures [4], the deeper or shallower MVGG architectures have not been more successful.

Finally, a new dataset that can be used in artificial intelligence researches has been prepared and a CNN architecture suggested that can be used in sign language image classification has been presented.

## 5 Conclusion and Future Works

Our dataset that can help to create machine learning algorithms which need to large and varied data to be successful. To get more successful results on use of the dataset, some data augmentation methods can be applied on our dataset to provide more success with CNN architectures. The number of series can be increased by rotating the images a few degrees from the center point clockwise or counterclockwise, it should always be noted that some sign language movements can have different meanings at different angles. Another method of data augmentation is adding random noise in a way that prevents images from losing data. These kind of data augmentation methods will increase the model success by increasing the number of data, and will make it more similar to daily use in future products. When these methods used during the training of the proposed architecture, 99% success rate was reached with the test dataset at the end of 15 training epochs.

Using this research, new algorithms that are more suitable for the problem area can be developed and real-time sign language translation products can be prepared for the daily use of the hearing impaired persons and people around them.

**Acknowledgements**

Thank to Zeynep Dikle and all of volunteer students and teachers from Ayrancı Anadolu High School for their help in collecting data.

*References*

[1] Mitchell, R. E., Young, T. A., Bachleda, B. and Karchmer, M. A., "How Many People Use ASL in the United States? Why Estimates Need Updating", Sign Language Studies, Volume 6, Number 3, 2006.

[2] LeCun, Y., Bottou, L., Bengio, Y. and Haffner, P., (1998), "Gradient-based learning applied to document recognition", *http://yann.lecun.com/exdb/publis/pdf/lecun-01a.pdf* (Date of access: 05.12.2020)

[3] Gu, J., Wang, Z., Kuen, J., Ma, L., Shahroudy, A., Shuai, B., Liu, T., Wang, X. Wang, L., Wang, G., Cai, J. and Chen, T., (2017), "Recent Advances in Convolutional Neural Networks", *https://arxiv.org/pdf/1512.07108.pdf* (Date of access: 05.12.2020)

[4] Hosseini, H., Xiao, B., Jaiswal, M. and Poovendran, R., (2017), "On the Limitation of Convolutional Neural Networks in Recognizing Negative Images", *https://arxiv.org/pdf/1703.06857.pdf* (Date of access: 05.12.2020)

[5] Zeiler, M. D., (2012), "ADADELTA: An adaptive learning rate method", *https://arxiv.org/pdf/1212.5701.pdf* (Date of access: 05.12.2020)
5